%% file: samplepaper.tex
\documentclass[runningheads]{llncs}
\usepackage{amsmath}
\usepackage{multirow}
\usepackage{multicol}
\usepackage{cite}
\usepackage{amssymb,amsfonts}
\usepackage{mathtools}
\usepackage{setspace}
\usepackage{algorithmic}
\usepackage{subcaption}
\usepackage{textcomp}
\usepackage{xcolor}
\usepackage{hyperref}
\usepackage{cleveref}
\usepackage{array}

\usepackage{graphicx}
\usepackage{amsmath}
\usepackage{kotex}	

\makeatletter
\newcommand{\printfnsymbol}[1]{%
  \textsuperscript{\@fnsymbol{#1}}%
}
\makeatother

\begin{document}
\title{DPRK-BERT: The Supreme Language Model}
\author{YEOJOO JEON\inst{1}\thanks{equal contribution} \and
Arda Akdemir\inst{2}\printfnsymbol{1}}
\authorrunning{Akdemir and Jeon}
\institute{
Graduate Schools for Law and Politics, University of Tokyo, Japan \and 
Institute of Medical Sciences, University of Tokyo, Japan \\}
\maketitle              % typeset the header of the contribution
\begin{abstract}
Deep language models have achieved remarkable success in the NLP domain. The standard way to train a deep language model is to employ unsupervised learning from scratch on a large unlabeled corpus. However, such large corpora are only available for widely-adopted and high-resource languages and domains. This study presents the first deep language model, DPRK-BERT, for the DPRK language. We achieve this by compiling the first unlabeled corpus for the DPRK language and fine-tuning a preexisting the ROK language model. We compare the proposed model with existing approaches and show significant improvements on two DPRK datasets. We also present a cross-lingual version of this model which yields better generalization across the two Korean languages. Finally, we provide various NLP tools related to the DPRK language that would foster future research.

\end{abstract}

\section{Introduction}

 Representation learning has been a major topic of interest in the NLP domain for many years~\cite{brown1992class}. The seminal work of Mikolov et al.~\cite{mikolov2013distributed}, Word2Vec, showed the success of learning fixed-size vector representations for each word by training neural networks on large unlabeled corpora. These representations go beyond the surface form and capture the semantic similarities between words. Despite its relative success in languages such as English, this word-level approach is less suitable for languages with rich morphology, such as the Korean languages. In such languages, the semantic information is retained in various morphemes, and a subword-level approach is necessary to capture this information. Rare words are also much more common in such languages, and word-level approaches fail to capture the characteristics of such words. Besides, such non-contextual approaches represent each word independent of its context, which causes a big problem for languages where there are many polysemous words such as the Korean language~\cite{kim2020korean}. More recently, deep learning-based pre-trained language models (PLMs) achieved remarkable success and became ubiquitous in the NLP domain~\cite{radford2018improving,peters2018deep,devlin2019bert}. Specifically, deep subword contextual representations such as the BERT embeddings~\cite{devlin2019bert} helped mitigate the above limitations of the previous approaches and achieved state-of-the-art results across many domains and languages. In fact, these languages models largely transformed the conventional way of training neural networks. Instead of training a specialized neural network from scratch on large task-specific labeled datasets, pre-trained language models are fine-tuned on significantly smaller datasets for each task and achieved better results. For example, language-specific models such as camemBERT~\cite{martin2019camembert} for French, outperform their counterparts while requiring minimal task-specific data. The standard way to obtain such a language model is to initialize the model weights randomly and train it on a large corpus in an unsupervised manner from scratch. However, such a large corpus is not available for many less-resourced languages. An extreme example is the Democratic People's Republic of Korea (DPRK) language. Due to the current circumstances, there are almost no digital resources available online for the DPRK Korean language, making it very difficult, if not impossible, to develop a DPRK PLM using the standard approach.

\begin{table}[ht]
    \centering
        \caption{Examples of distinctive differences between the two Korean Languages~\cite{kwon2015})}
    \label{tab:dprkvskor-summary}
    
   \resizebox{\textwidth}{!}{\begin{tabular}{c|c|c|c}\hline
     \textbf{Description}& \textbf{DPRK}& \textbf{ROK}& \textbf{English}\\\hline
      The middle ㅅ rule  &나무잎 &나뭇잎 & leaf \\
      Initial sound rule  &로동 &노동 & labor \\
      Different word usage &동무 & 동무 & revolutionary comrade(DPRK)friend(the ROK)\\
     DPRK-specific ideological terms &로동영웅& - & labor hero \\
     White-space usage difference &갈바를\_알수\_없다&갈 바를\_알\_수\_없다& don't know how to go\\\hline 
    \end{tabular}}

\end{table}

There are many publicly available labeled datasets for Republic of Korea (ROK) language~\cite{jang2013kosac,lim2019korquad1,ham2020kornli}. Moreover, obtaining large unlabeled corpora through web crawling or through Wikipedia dumps is effortless in the case of the ROK. These resources have greatly fostered Korean NLP research. For example there are various high-performing deep PLMs for the ROK language such as KoBERT~\cite{korbert2019}, KorBERT~\cite{sktbrain}, and KR-BERT~\cite{lee2020kr}. These models are applied to various downstream tasks to achieve state-of-the-art results~\cite{lee2020kr}. Yet, the training data of these models do not include DPRK-specific text. Over the years, the Korean language used in the countries diverged significantly from each other. Table~\ref{tab:dprkvskor-summary} highlights some of the important differences in the written Korean language between the two countries. As a result of these differences, we can not directly apply the RoK PLMs to analyze the DPRK data.

%There are many differences in the use of the common vocabulary, there are many borrow

%These models are unable to understand the intentions of using specific words about political agendas as these words almost never occur in RoK Korean text. Besides systematic differences in the written text such as differences in white-space usage for noun phrases and auxiliary and difference in spelling of various Sino-Korean (SK) words limit the applicability of RoK language models directly to DPRK texts.  

The current NLP research on the DPRK language is highly limited in volume and scope because of a lack of NLP tools and resources. Most NLP research on the language tries to mitigate this issue either by relying on domain experts' manual readings of the DPRK articles or by only focusing on the English articles provided by the KCNA(Korean Central News Agency). The use of NLP methods in all these studies is limited to very conventional keyword-based approaches.

Increasing the available NLP tools and resources, and developing a high-quality DPRK language model is extremely important to enable better interpretations of the DPRK messages and prevent misperceptions of the DPRK's political agenda. Besides, there is an increasing amount of the DPRK's refugees worldwide and especially in the ROK~\cite{green2013now}. A genuinely inclusive AI requires developing tools for all communities regardless of its difficulty due to the scarcity of resource.

% - Avoid misunderstanding and improve the interpretation of the DPRK messages.
% - Prevent error to decipher the DPRK’s perceptions/signals on political agendas.
% - Truly inclusive AI is only possible when all communities are represented equally well. There are many North Korean refugees in South Korean currently~\cite{green2013now}.

% - There are many urgent political issues related to DPRK:

% Nuclear ambitions/Non-proliferation
% Unification issues
% Japanese abduction issues
% China-DPRK relations
% and this suggest the importance of better interpreting the DPRK signals. 

The above observations motivated us to develop the first deep language model, DPRK-BERT, for the DPRK language. To this end, we compiled the first large DPRK corpus based on the KCNA released Rodong newspaper articles. We used the Transformer-based~\cite{vaswani2017attention} BERT model~\cite{devlin2019bert} as the deep learning model. We initialize the model weights using the state-of-the-art RoK Korean PLM, KR-BERT~\cite{lee2020kr} and continue its pretraining on the Rodong corpus using masked language model (MLM) training. Evaluated on two different DPRK Korean datasets, the proposed LM significantly outperformed all other Korean LMs. The pretrained DPRK-BERT model as well as the compiled DPRK Korean dataset are publicly available.~\footnote{https://github.com/ardakdemir/DPRK-BERT}.

Continuing the pre-training of PLMs for task-specific purposes is a recurring theme in the NLP research. Domain adaptive pre-training (DAPT) and even task adaptive pre-training (TAPT) are shown to yield performance improvements on the downstream tasks~\cite{gururangan2020don}. Given a downstream target task $\mathcal{T}$ with a domain $\mathcal{D}_\mathcal{T}$, DAPT refers to continuing the training of a PLM on an unlabeled dataset from the same domain $\mathcal{D}_\mathcal{T}$, before applying the PLM on task $\mathcal{T}$. Similarly, TAPT continues the pretraining on the unlabeled version of the dataset for task  $\mathcal{T}$. Intuitively, both approaches exposes the PLM to data with a similar distribution to that of the target task, that allows tailoring the weights of the PLM to the downstream task. DAPT is especially useful when the size of the available unlabeled data is big compared to the labeled dataset of task $\mathcal{T}$. 

Yet, naively continuing the pretraining on a dataset with unique data distribution and language usage, such as the DPRK news articles, results in catastrophic forgetting~\cite{kirkpatrick2017overcoming}. Catastrophic forgetting refers to forgetting the well-learned information obtained on an old task when learning a new task. In the case of neural networks, catastrophic forgetting occurs with the uncontrolled altering of the weights of the network when learning a new task.  

\begin{figure}
    \centering
    \includegraphics[width=\textwidth]{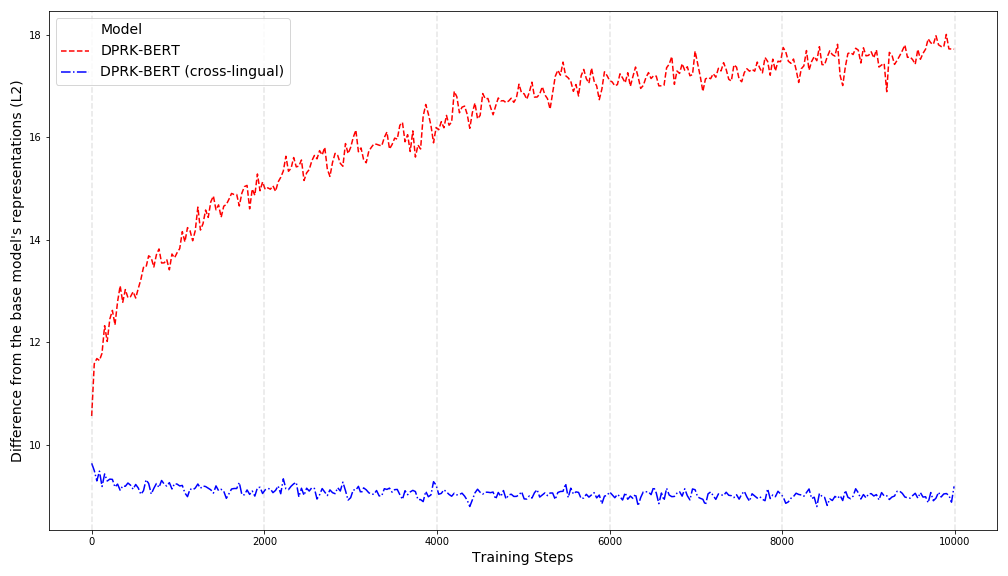}
    \caption[Illustration for weight stray during continued pre-training. ]{L2 norm between the representations of the new model and the base model during training (the lower the better for achieving a cross-lingual model).}
    \label{fig:catastro-korean}
\end{figure}

In order to obtain a cross-lingual Korean PLM that maintains good performance on RoK datasets while being trained on the DPRK datasets, we introduce a cross-lingual regularization term. By adding this simple regularization into the fine-tuning stage of the PLM we obtain significantly better generalization over two Korean datasets. Figure~\ref{fig:catastro-korean} illustrates
the difference between uncontrolled and controlled altering of the KR-BERT model's weights during the continued pretraining. DPRK-BERT is trained without cross-lingual loss term which we explain later in detail.

Our main contributions in this work are as follows:

\begin{enumerate}
    \item We present the first DPRK deep language model by fine-tuning the KR-BERT model on a large unlabeled DPRK dataset.
    \item We improve on this model by introducing a cross-lingual regularization term which achieves significantly better generalization on the two Korean languages than all other approaches. We refer to this cross-lingual model as CL-DPRK-BERT. 
    \item We provide the first DPRK dataset publicly available in electronic format.
    \item We give a comparison of four different deep LMs on two different DPRK  datasets.
    \item We build several NLP tools for future research on DPRK such as a web scraper for Rodong news articles.
\end{enumerate}

\section{Related Work}

Until now, there have been very few works on applying NLP methods to DPRK data. Besides, the existing works mostly rely on traditional NLP methods such as keyword-based analyses and are also highly dependent on the knowledge of the domain experts. Below we briefly explain the previous research in this area.

Park et al.~\cite{park2015automated} analyzes the new year speeches between 1946 and 2015. For their analysis, they used the ROK Word2Vec embeddings for analyzing the word cluster. Due to the unavailability of a large corpus and a DPRK-specific representation model, they resorted to using the ROK embeddings that have several limitations due to the differences between the two languages. They also analyzed the sentiment of the addresses about certain entities such as the US, the ROK, and Japan. To this end, domain experts manually reviewed the documents to capture a list of keywords to decipher the trends of the DPRK's perception of these entities. Even though their findings provide crucial knowledge on the DPRK, we believe that going beyond keyword-based methods and relieving the burden for human experts is critical for future research. 

Park~\cite{park2019discovering} conducted an analysis over Kim Jong Un’s inaugural speeches from 2017 to 2019. They analyzed the co-occurrence of terms to find out the key and central topics from the speeches. Their work is mostly based on word counts and the co-occurrences of words. Additionally, it does not utilize machine learning methods or language models. 

Whang et al.~\cite{whang2018detecting} used English articles provided by KCNA from 1997-2013 to find patterns about the DPRK's military provocations. They labeled 1,624 English articles with \texttt{Threat} and \texttt{Non-Threat} labels, and used this dataset to train a predictive machine learning model. They provide a useful model for predicting DPRK-related threats but, it is only applicable to English data.

The limitation of the KoNLP for DPRK research is apparent in Lee et al.~\cite{Lee2021UNSanc}. They analyzed the official statements issued by the DPRK authorities through the KCNA in response to the UN Security Council's resolutions on sanctions against DPRK. However, they only utilized the English statements by DPRK. They admit that it would be beneficial to analyze the Korean statements. However, they argue because the current KoNLP is based on the ROK language, it has a problem of relevance in lexical part-of-speech classification and morpheme analysis.

\section{Linguistic Differences between the two Korean Languages}

Since the Division of Korea around 70 years ago, the language used in the two countries diverged significantly~\cite{ramsey2012korean}. Below is a list of the major differences in the two languages~\footnote{The differences between the two languages were taken from Kwon~\cite{kwon2015} Refer to the paper for more examples.} (see Table~~\ref{tab:dprkvskor} for examples):

\begin{enumerate}

    \item There are a significant number of words that have different meanings and usages in the two languages.  
    
%     \begin{table}[ht]
%     \centering
%         \caption{Examples of words that are spelled the same but have different meanings (examples are taken from~\cite{kwon2015})}
%     \label{tab:dprkvskor}
    
%   \resizebox{\textwidth}{!}{\begin{tabular}{c|l|l}
%      Word & Meaning in the DPRK& in the ROK\\\hline
%       동무 & revolutionary comrade & friend \\\hline
%       \multirow{3}{*}{어버이} & who provides the most noble political & \multirow{3}{*}{parents}\\
%       &  life and gives love and consideration &\\
%       & more than its biological parents&\\\hline 
%     \end{tabular}}

% \end{table}

    \item There are many borrowed words from Western languages in the ROK language that the DPRK language doesn't use. The ROK words sound similar to the English equivalents but, the DPRK created words by combining existing Korean words (see Table~\ref{tab:dprkvskor-2}).

% \begin{table}[ht]
%     \centering
%         \caption{Examples of the ROK words with foreign origins that are not used in the DPRK language(examples are taken from DPRK information portal site)}
%     \label{tab:dprkvskor}
    
%   \resizebox{\textwidth}{!}{\begin{tabular}{c|c|c|c}
%      Meaning & DPRK& ROK & Description\\\hline
%       Goalkeeper & 문지기 & 골키퍼 & The ROK words sound like the English equivalents but the DPRK created words from Korean\\
%       Knockdown & 맞아넘어지기 & 녹다운 \\
%       Net & 배구그물 & 네트 \\
%       Dynamite & 남포약 & 다이너마이트\\\hline 
%     \end{tabular}}

% \end{table}

   \item The two languages spell foreign countries' names differently. 
   
% \begin{table}[ht]
%     \centering
%         \caption{Examples of foreign countries names(examples are taken from DPRK information portal site)}
%     \label{tab:dprkvskor}
    
%   \resizebox{\textwidth}{!}{\begin{tabular}{c|c|c}
%      Meaning & DPRK& ROK\\\hline
%       Nigeria & 나이제리아 & 나이지리아\\
%       Netherlands & 네데를란드 & 네덜란드 \\
%       Denmark & 단마르크 & 덴마크 \\\hline 
%     \end{tabular}}

% \end{table}

    \item The DPRK has newly generated words, especially derived from its political ideology, that almost never occur in the ROK language or are used in different manners.
    
% \begin{table}[ht]
%     \centering
%         \caption{Examples of ideological vocabulary in the DPRK language (examples are taken from~\cite{kwon2015})}
%     \label{tab:dprkvskor}
    
%   \resizebox{\textwidth}{!}{\begin{tabular}{c|c|c}
%      DPRK& ROK& Meaning\\\hline
%       원쑤 & - & The DPRK language distinguish '원수(the honorific way of calling the supreme leaders)' and '원쑤(the enemy)' but, the ROK doesn't.\\
%       공훈리발사 & - & A barber who made a remarkable contribution for the country. The ROK doesn't have this concept. \\
%     \hline 
%     \end{tabular}}

% \end{table}

%Also, I referred to this webpage for the description of the first word in the above table. Could you put this reference to the table as well?
%http://www.tufs.ac.jp/ts/personal/choes/korean/nanboku/Snanboku.html

    \item There are distinct variations in the spelling of certain syllables (in Sino-Korean words).

% \begin{table}[ht]
%     \centering
%         \caption{Examples of the difference in initial spelling(examples are taken from~\cite{kwon2015})}
%     \label{tab:dprkvskor}
    
%   \resizebox{\textwidth}{!}{\begin{tabular}{c|c|c}
%      Meaning & DPRK& ROK\\\hline
%       Labor & 로동 & 노동\\
%       Woman & 녀자 & 여자 \\\hline 
%     \end{tabular}}

% \end{table}
    
    \item The spacing rules for noun phrases and auxiliary words are different in the two languages.
    
% \begin{table}[ht]
%     \centering
%         \caption{Examples that show different spacing rules(examples are taken from~\cite{kwon2015})}
%     \label{tab:dprkvskor}
    
%   \resizebox{\textwidth}{!}{\begin{tabular}{c|c|c}
%      Meaning & DPRK& ROK\\\hline
%       our country & 우리나라 & 우리 나라\\
%       knowing & 아는것 & 아는것 \\
%       important thing & 중요한것 &중요한 것\\
%       going(present progressive) & 가고있다 & 가고 있다\\\hline 
%     \end{tabular}}

% \end{table}

\end{enumerate}

\begin{table}[h]

     \caption{Examples for all important differences in the Korean Languages}
     \label{tab:dprkvskor}
    \begin{subtable}[h]{0.45\textwidth}
       \centering
        \caption{Same spelling with different meaning~\cite{kwon2015}}
    \label{tab:dprkvskor-1}
   \resizebox{\textwidth}{!}{\begin{tabular}{c|l|l}
     Word & DPRK& ROK\\\hline
      동무 & revolutionary comrade & friend \\\hline
      \multirow{3}{*}{어버이} & who provides the most noble political & \multirow{3}{*}{parents}\\
      &  life and gives love and consideration &\\
      & more than its biological parents&\\\hline 
    \end{tabular}}
    \end{subtable}
    \hspace{1cm}
    \begin{subtable}[h]{0.45\textwidth}
        \centering
        \caption{Foreign words in the ROK language~\cite{dprkinfoportal}}
    \label{tab:dprkvskor-2}
\begin{tabular}{c|c|c}
     Meaning & DPRK& ROK\\\hline
      Goalkeeper & 문지기 & 골키퍼\\
      Knockdown & 맞아넘어지기 & 녹다운 \\
      Net & 배구그물 & 네트 \\
      Doughnut & 가락지빵 & 도넛\\\hline 
    \end{tabular}
     \end{subtable}
     
      \begin{subtable}[h]{0.5\textwidth}
       \centering
        \caption{Foreign country names~\cite{dprkinfoportal}}
    \label{tab:dprkvskor-3}
\begin{tabular}{c|c|c}
     Meaning & DPRK& ROK\\\hline
      Nigeria & 나이제리아 & 나이지리아\\
      Netherlands & 네데를란드 & 네덜란드 \\
      Denmark & 단마르크 & 덴마크 \\\hline 
    \end{tabular}
     \end{subtable}
      \begin{subtable}[h]{0.5\textwidth}
      \centering
        \caption{DPRK ideological terms~\cite{kwon2015}}
    \label{tab:dprkvskor-5}
   \resizebox{\textwidth}{!}{\begin{tabular}{c|c|l}
     DPRK& ROK& Meaning\\\hline
      \multirow{3}{*}{원쑤} & \multirow{3}{*}{-} & The DPRK language distinguish \\
      &&'원수(the honorific way of calling the supreme leaders)' \\
      &&and '원쑤(the enemy)' but, the ROK doesn't.\cite{langdifftufs}\\\hline
       \multirow{3}{*}{공훈리발사} &  \multirow{3}{*}{-} & A barber who made  \\
       &&a remarkable contribution for the country.\\
       &&The ROK doesn't have the concept. \\
    \hline 
    \end{tabular}}
     \end{subtable}
    
      \begin{subtable}[h]{0.45\textwidth}
       \centering
        \caption{Initial spelling diffference~\cite{kwon2015})}
    \label{tab:dprkvskor-6}
  \begin{tabular}{c|c|c}

     Meaning & DPRK& ROK\\\hline
      Labor & 로동 & 노동\\
      Woman & 녀자 & 여자 \\\hline 
    \end{tabular}
     \end{subtable}
     \hspace{1cm}
     \begin{subtable}[h]{0.45\textwidth}
      \centering
        \caption{White-space difference~\cite{kwon2015})}
    \label{tab:dprkvskor-7}
   \resizebox{\textwidth}{!}{\begin{tabular}{c|c|c}
     Meaning & DPRK& ROK\\\hline
      our country & 우리나라 & 우리 나라\\
      knowing & 아는것 & 아는 것 \\
      important thing & 중요한것 &중요한 것\\
      going(present progressive) & 가고있다 & 가고 있다\\\hline 
    \end{tabular}}
     \end{subtable}
     
\end{table}

These systematic differences between the two Korean suggest that 1) applying the PLMs trained on the RoK Korean datasets directly on DPRK data would fail to generate meaningful representations for the DPRK-specific phrases and sentences, and 2) continuing the fine-tuning of the RoK PLMs on the DPRK data without regularization would result in the catastrophic forgetting of the learned representations for the RoK Korean.

\section{Datasets}
\label{dprk-datasets}
There are no publicly available NLP datasets specifically for the DPRK language. In this section, we explain the datasets we compiled for the DPRK language as well as the RoK Korean dataset we used for evaluating the proposed models.

\subsection{Rodong News articles}

The only North Korean online data provider is the Korean Central News Agency (KCNA). KCNA is the only news agency in North Korea and provides content for both domestic and foreign audience~\cite{shrivastava2007news}. KCNA releases daily news articles under Rodong Sinmun (Worker's Newspaper in English) toward its domestic audience. Every day, around 30 news articles from Rodong are made available online under the official website.~\footnote{http://www.rodong.rep.kp/ko/index.php} The online version is continued from the beginning of 2018 and there are approximately 50 thousand articles including . 

\subsubsection{Web scraper for Rodong News}

The official Rodong website supports searching. However, there are no mechanism to download the news articles in text format. Considering the large size of the database,
this makes it practically impossible to obtain the whole dataset manually. For this reason, we developed a web scraper for automatically parsing the official Rodong website to download all news articles in text format. 

As accessing this dataset does not require any permission online, at least where we are currently based, we can provide it as a single file with the web scraper software to foster future research upon request.~\footnote{Note that sharing it online as an open-source may have some legal issues. We will decide how to make this sharing process possible in consulting with the relevant parties.} We recommend using the software, as the data keeps growing every day with the forthcoming news articles.

\subsection{New Year Addresses}

The second dataset we used in our experiments is the Inaugural New Year Addresses of the Supreme Leader(s). Each year, the Supreme Leader of DPRK addresses its own citizens and these speeches contain valuable information about the political intentions and the agenda of Pyongyang~\cite{park2015automated}. The new year addresses are available for every year from 1946 to 2019 except 1957, when it was not released due to its domestic political issues~\cite{newyearspeechmk}. We used the annual addresses in text format~\footnote{For data formation, we follow the criteria provided in Part et al. ~\cite{park2015automated} By referring to ~\cite{peaceproblem1997}, they coded 1) messages of congratulations, 2) speeches, 3) New Year's editorials,  4) joint editorials of the three major newspapers (Rodong Sinmun(로동신문), Joson Inmingun(조선인민군), and Chongnyon Jonwi(청년전위)) as new year addresses. Our dataset includes Kim Il-sung's New Year's Address (1946-1950, 1954-1956, 1957-1965, 1969, 1971-1994), the New Year's Editorial by the Rodong Sinmun during Kim Il-sung's regime (1966-1968, 1970), and New Year's Messages to the North Korean Army (1951-1953), New Year's Editorial during Kim Jong-il's regime (1995-2011), Kim Jong-un's New Year's Editorial (2012), and oral New Year's Address (2013-2019).~\cite{newyearadd1946-2019}; The DPRK has not been issuing its annual new year addresses since 2020 but substituting them with the Report of the Plenary Meeting of the Central Committee of the Workers party in 2020 and a letter written by Kim Jong-un in 2021.~\cite{newyearspeechmk}}. More details about each dataset are given in Table~\ref{tab:datasets}.

\begin{table}[ht]
    \centering
        \caption{Details about each DPRK dataset}
    \label{tab:datasets}
    \begin{tabular}{c|c|c}
    &New Year Speeches & Rodong Sinmun \\\hline
       Number of Documents  & 73 &27,401  \\
       Number of Sentences  &5,709& 471,417\\
       Date Span &1946-2019& 2018-now\\\hline
       
    \end{tabular}

\end{table}

\subsection{RoK Korean Dataset.}

In order to evaluate the cross-lingual capabilities of the models, we also experimented with RoK Korean data. We used the Korean Natural Language Inference (KorNLI) dataset~\cite{ham2020kornli}. KorNLI contains a total of 950,354 examples annotated for the NLI task. The majority of this data (942,854 examples) is obtained by using machine translation to translate several English NLI datasets. These machine translated examples are used for the training split. For the development and the test splits, the authors used human translations of the same datasets. Table~\ref{tab:kornli-details} gives the details of the KorNLI dataset. For our evaluation purposes, we only utilized the development and the test splits of this datasets. An NLI example consists of two parts: premise and hypothesis. As we used this data for the masked language modeling task, we simply concatenated the premise and the hypothesis sentences of each example.

\begin{table}[ht]
    \centering
    \caption{Details of the KorNLI dataset.}
    \label{tab:kornli-details}
    \begin{tabular}{c|cccc}
 &	Total&	Train&	Dev.&	Test\\\hline
Translated by&	-&	Machine	&Human&	Human\\\hline
\# Examples&	950,354	&942,854&	2,490&	5,010\\
Avg. \# words (premise)&	13.6&	13.6&	13.0&	13.1\\
Avg. \# words (hypothesis)&	7.1	&7.2&	6.8 &	6.8\\\hline
    \end{tabular}
\end{table}

\subsection{Mapping novel syllables}

Both Korean languages are written using the Hangul alphabet. In the Hangul alphabet individual characters are combined to form syllables and syllables constitute words. As two languages significantly diverged from each other over the last 70 years, there are spelling differences and unique words in both languages. We observed that the KR-BERT tokenizer's vocabulary did not contain 52 syllables that occurred in the Rodong News articles. Some of these syllables occur very frequently. For example, the syllable ``돐'' occurs 5,449 times and is not defined in the KR-BERT vocabulary. 

Having a large number undefined syllables causes a significant problem for modeling a language. This is because all the words that contain an unknown syllable is tokenized as an unknown word \texttt{[UNK]} by the BERT tokenizer. Below are examples KR-BERT tokenization for DPRK sentences,

\begin{center}
\texttt{Input}:  ... 우리 천만군민으로 하여금 한걸음의 주저도 한치의 에돎도 없이 경애하는 원수님을 따라\\
\texttt{Tokenized}: ... 우리 천 \#\#만 \#\#군 \#\#민 \#\#으로 하여 \#\#금 한 \#\#걸 \#\#음 \#\#의 주 \#\#저 \#\#도 한 \#\#치 \#\#의 [UNK] 없이 경 \#\#애 \#\#하는 원 \#\#수 \#\#님 \#\#을 따라\\
\texttt{Unknown token}:  돎\\
\end{center}
The whole word ``에돎도'' is returned as \texttt{[UNK] } as it contains the unknown syllable \texttt{돎}. The ROK language doesn't have such word.

\begin{center}
\texttt{Input}:  ... 일본 단마르크 도이췰란드 로므니아 로씨야 벌가리아 벨라루씨 스위스 스웨리예 슬로베니아 체스꼬\\
\texttt{Tokenized}: ... 일본 단 \#\#마 \#\#르크 [UNK] 로 \#\#므 \#\#니아 로 \#\#씨 \#\#야 벌 \#\#가 \#\#리아 벨 \#\#라 \#\#루 \#\#씨 스위스 스 \#\#웨 \#\#리 \#\#예 슬 \#\#로 \#\#베 \#\#니아 체 \#\#스 \#\#꼬 \\
\texttt{Unknown token}:  췰\\
\end{center}
The whole word ``도이췰란드(Germany)'' is returned as \texttt{[UNK] } as it contains the unknown syllable \texttt{췰}. The ROK uses ``독일'' for Germany.
\vspace{1cm}

% \begin{center}
% \texttt{Input}:  ... 앙드레 로헤껠레 깔론다는 다음과 같이 말하였다\\
% \texttt{Tokenized}: ... 앙 \#\#드 \#\#레 [UNK] 깔 \#\#론 \#\#다 \#\#는 다음과 같이 말 \#\#하였다\\
% \texttt{Unknown token}:  껠\\
% \end{center}
% The whole word ``로헤껠레'' is returned as \texttt{[UNK] } as it contains the unknown syllable \texttt{껠}. The two countries' different rules of spelling foreign words cause this tokenization failure. 

To mitigate this issue, a native Korean speaker with experience in the DPRK Korean characteristics defined a mapping between DPRK-specific new syllables and syllables available in KR-BERT vocabulary. Table~\ref{tab:syl-map} shows some examples for the syllable mapping.

\begin{table}[ht]
    \centering
        \caption{Examples for the syllable mapping.}
    \label{tab:syl-map}
    \begin{tabular}{c|c}\hline
      DPRK&ROK\\\hline
돐&주년\\
윁&베트\\
췰&칠\\
꾜&쿄\\
뙈&떼\\
곬&골\\\hline
    \end{tabular}

\end{table}

\section{Methodology}

\subsection{BERT Language Model}

\begin{figure}
    \centering
    \includegraphics[width=\textwidth]{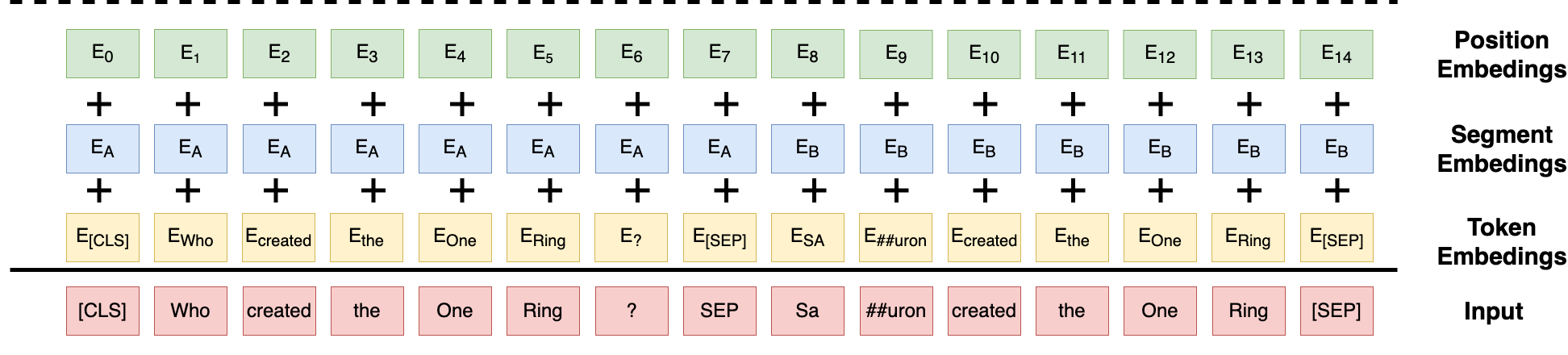}
    \caption{An illustration for the BERT model's input.}.
    \label{fig:bert-illus}
\end{figure}

In this study, we used \textbf{B}idirectional \textbf{E}ncoder \textbf{R}epresentations from \textbf{T}ransformers (BERT in short)~\cite{devlin2019bert} as the deep language model. BERT is a bidirectional transformer encoder that is trained using the masked language model (MLM) and the next sentence prediction (NSP) tasks. 
BERT's architecture is identical to the Transformer-encoder architecture~\cite{vaswani2017attention}. BERT uses a WordPiece~\cite{wu2016google} tokenizer to represent an input sequence as a list of subword tokens. Given a training corpus, preferably a huge corpus that is representative of the true usage of the words, and a vocabulary size $v$, WordPiece uses an algorithm similar to the byte-pair encoding (BPE)~\cite{gage1994new} algorithm to determine a subword vocabulary $V$ based on the frequency of occurrence of all characters and character sequences. Frequent words like stop words are included in the tokenizer vocabulary. Other words are represented as the concatenation of subword tokens inside the vocabulary. Non-starting subword tokens are preceded by ``\#\#'' to denote they do not start a new word. As illustrated in Figure~\ref{fig:bert-illus}, BERT's input is the concatenation of three embeddings: token embedding, position embedding, and segment embedding. Position embeddings are necessary to encode the positional information as the attention mechanism itself is invariant to the relative positions of the input words. Segment embeddings are used when the input consists of a sentence pair instead of a single sentence, as in the case for question answering task where the input is made up of a question followed by a passage.

Following the previous work, we used the masked language modeling (MLM) for pretraining the BERT model on the unsupervised Rodong News articles dataset. In the MLM task, the model is given a sentence where certain words are replaced with a \texttt{[MASK]} token and the model tries to predict the correct words from its vocabulary $\mathcal{V}$ of size $\mathcal{M}$. 

Formally, given a dataset $\mathcal{D}=\{(\mathcal{Y}_1,\mathcal{X}_1),...,(\mathcal{Y}_N,\mathcal{X}_N)\}$ the objective function to minimize is the average negative log likelihood over all masked input sentences:
\begin{align*}
    NLL(\mathcal{Y},\mathcal{X},\Theta) &= - \log p(\mathcal{Y}|\mathcal{X},\Theta)\\
    \tilde{\Theta} &= \arg \max_{\Theta} \sum_{(\mathcal{Y},\mathcal{X})\in \mathcal{D}}  NLL(\mathcal{Y},\mathcal{X},\Theta), \\
\end{align*}

where $\mathcal{Y}_1$ and $\mathcal{X}_1$ refers to the correct word sequence and the input to the language model, respectively. Similarly, log-\textit{perplexity} is defined as the average negative log-likelihood of all inputs. Intuitively, it is a measure of how much the model is surprised to see the correct sequence. 
\begin{align*}
    \text{log-}perplexity(\mathcal{D},\Theta) &= \dfrac{1}{N}\sum_{(\mathcal{Y},\mathcal{X})\in \mathcal{D}} NLL(\mathcal{Y},\mathcal{X},\Theta).\\
\end{align*}

Deep learning models such as BERT, require large amount of data in order obtain meaningful representations as they have billions of parameters to tune. The Rodong News article dataset does not have enough coverage to learn a language model from scratch. For this reason, instead of training the BERT model from scratch, we initialized the model weights using the state-of-the-art RoK Korean LM, KR-BERT. In a way, we continue the pretraining of the KR-BERT model, on the DPRK-specific data.

\subsection{Training}

We trained the DPRK-BERT model by using the MLM task and by initializing its weights with the publicly available KR-BERT checkpoint. For the base DPRK-BERT model, we used the NLL loss over the masked tokens for updating the weights of the deep learning model without the regularization term. Given an input sentence $S_i$ with masked tokens, let $\mathcal{C}_i$ represent the set of masked indices. For each index $m \in \mathcal{C}_i$, let $y_m$ represent the unmasked token at position $m$ in $S_i$. Then, the MLM loss on $S_i$ is as follows:

% [Define cross-entropy loss for a single sentence here]

$$
\mathcal{L}_{i}^{MLM} =  \sum_{m \in \mathcal{C}_i}  - \log p(S_m = y_m  | S ,\Theta , m).
$$

When possible, we used the default hyperparameters for the model architecture and for training. Table~\ref{tab:hypers} shows the important hyperparameters of the proposed model. 

\begin{table}[ht]
    \centering
        \caption{Hyperparameters of the DPRK-BERT model.}
    \label{tab:hypers}
    \begin{tabular}{c|c}
        Parameter Name & Value  \\\hline
          attention probs dropout prob& 0.1\\ 
    hidden act& gelu\\ 
    hidden dropout prob& 0.1\\ 
    hidden size& 768\\ 
    initializer range& 0.02\\ 
    intermediate size& 3072\\ 
    max position embeddings& 512\\ 
    num attention heads& 12\\ 
    num hidden layers& 12\\ 
    type vocab size& 2\\ 
    vocab size& 16424\\ 
    mask probability & 0.15\\
    number of epochs & 20\\
    layer norm eps& 1e-12\\\hline
    \end{tabular}

\end{table}

\subsection{Cross-lingual Regularization}

Catastrophic forgetting is an important challenge when sequentially learning multiple tasks~\cite{kirkpatrick2017overcoming}. Training on a large amount of DPRK data would result in overfitting and catastrophic forgetting of the well-learned RoK Korean representations. For this reason, we introduce a cross-lingual regularization term that penalizes the model for straying from the initial KR-BERT representations. We name this model CL-DPRK-BERT to refer that it is a cross-lingual model.

% [Define cross-entropy + regularization loss for a single sentence here]
Given a sentence $S_i$ consisting of words $x_j \in S_i$, the total loss $\mathcal{L}_i$ is the combination of MLM loss and the cross-lingual regularization term:

\begin{align*}
    R_i &=  \sum_{j=1}^{|S_i|} \left\lVert f_0(x_j) - f(x_j) \right\rVert^2\\
\mathcal{L}_i &= \mathcal{L}_{i}^{MLM} + \lambda R_i,
\end{align*}

where $f_0(.)$ is the initial KR-BERT model,  $f_(.)$ is the current LM, and $\lambda$ is the weight for the regularization term. $\lambda$ is tuned with a hyper-parameter search. It is common to apply regularization to prevent the newly learned parameters from deviating too much from the initial ones. A standard way to achieve this is to define an $l_2$ regularization on the difference between the model's new weights and the
initial weights~\cite{wiese2017neural}. In addition to the $l_2$ regularizer, Wiese et al.~\cite{wiese2017neural} also use a forgetting cost to prevent catastrophic forgetting. Their loss is defined as the cross-entropy loss between the new predictions and the initial model's predictions. In contrast, we define a different cost value based on the difference between the representations. Our cross-lingual regularization term resembles the regularizer introduced by Cao et al.~\cite{cao2019multilingual}. During the training of cross-lingual word embeddings using a word-aligned bilingual dataset, they penalize the model if the new representations stray from the initial ones. They only add the regularization term on the aligned words, whereas we apply it to all words in the sentence.

\section{Results}
In this section we report the results obtained on the three datasets introduced in Section~\ref{dprk-datasets}. First, we report results on the Rodong test set. Following that, we present the results on the KorNLI dataset and discuss the generalization of all language models. Finally, we further evaluate the proposed methods on the New Year Addresses. In the following text, DPRK-BERT refers to the DPRK LM trained without the cross-lingual regularization term and CL-DPRK-BERT refers to the one trained with the regularization term.

We compare our approach with three other deep learning models:
\begin{enumerate}
    \item KR-BERT~\cite{lee2020kr}: It is the state-of-the-art LM for the South Korean Language.
    \item KR-BERT-MEDIUM: KR-BERT's extended version which is trained using, in addition to the original dataset, legal texts crawled from the National Law Information Center and a Korean comment dataset. KR-BERT-MEDIUM is shown to outperform KR-BERT on the sentiment analysis task.~\footnote{https://github.com/snunlp/KR-BERT-MEDIUM}
    \item mBERT~\cite{devlin2019bert}: The multilingual BERT model trained on Wikipedia articles of 102 languages. Korean is also included in the training set however it is much less represented compared to the English and several other more resourced languages.

\end{enumerate}

We used \textit{Perplexity} and \textit{MLM Accuracy} metrics to evaluate the performance of each model. \textit{MLM Accuracy} is the rate of the correctly predicted masked tokens, the higher the better. For \textit{Perplexity}, a lower score denotes a better language model. We repeated evaluation on each dataset three times as evaluation involves randomness (tokens are masked randomly). 

\subsection{Rodong Sinmun Results}

First, we evaluate all four models on the validation split of the Rodong Sinmun dataset. The results for both log \textit{Perplexity} and \textit{MLM Accuracy} are given in Table~\ref{tab:mlm-rodongandkornli}. First, we see that the multilingual BERT model performs significantly worse than all other PLMs ($3.410$ average log perplexity and $37.730\%$ average MLM accuracy). This supports the previous research that illustrates the weaknesses of multilingual LMs in low-resource settings~\cite{pires2019multilingual,virtanen2019multilingual,lee2020kr}. However it must be noted that the vocabulary size of mBERT tokenizer is significantly larger than other LMs. A larger tokenizer vocabulary makes it more difficult to find the correct masked token, so mBERT has a disadvantage over other models. More importantly, we see that the DPRK-BERT model significantly outperforms all other approaches, $1.702$ improvement on perplexity and $28.8\%$ improvement on accuracy over the closest model (KR-BERT). DPRK-BERT achieves an average of $82.37\%$ MLM accuracy. An interesting observation is that the KR-BERT model significantly outperforms KR-BERT-MEDIUM on both metrics. This is slightly surprising as KR-BERT-MEDIUM training data is an extension of KR-BERT training data. This performance drop might be due to the inclusion of user-generated Korean Comment dataset, which has significantly different use of the Korean language compared to Rodong Sinmun articles, in the training corpus of KR-BERT-MEDIUM.

The last two rows show the models that are trained using the cross-lingual regularization term. We see that their performance is significantly higher than the other three LMs compared.  CL-DPRK-BERT with $\lambda=0.3$ outperforms the base model KR-BERT by more than $16\%$ in MLM accuracy.  

\begin{table}[ht]
    \centering
        \caption{MLM Results for all LMs  on Rodong (DPRK) and KorNLI (RoK) datasets. }
    \label{tab:mlm-rodongandkornli}
    \resizebox{\textwidth}{!}{\begin{tabular}{c|cc|cc|cc}\hline
    &\multicolumn{2}{c}{Rodong}  &\multicolumn{2}{c}{KorNLI}  &\multicolumn{2}{c}{Average}\\
Model&Perplexity&Accuracy&Perplexity&Accuracy&Perplexity&Accuracy\\\hline
mBERT&3.41&37.734&3.066&42.084&3.238&39.909\\
KR-BERT-MEDIUM&3.241&43.129&2.928&45.553&3.085&44.341\\
KR-BERT&2.505&53.57&2.536&51.426&2.52&52.498\\\hline
DPRK-BERT&0.803&82.367&5.465&25.788&3.134&54.078\\
CL-DPRK-BERT ($\lambda=0.2$)&1.495&68.429&2.903&45.317&\textbf{2.199}&56.873\\
CL-DPRK-BERT ($\lambda=0.3$)&1.448&69.716&2.989&45.067&2.218&\textbf{57.391}\\\hline
    \end{tabular}}
\end{table}

\subsection{KorNLI Results}

Next, we evaluated all models on the KorNLI dataset, which contains RoK Korean examples. The main motivation of this step is to compare the generalization of different PLMs on the Korean languages. As expected, the KR-BERT model achieves the best results on both evaluation metrics. KR-BERT achieves $51.426\%$ average MLM Accuracy on this dataset and outperforms all other approaches by more than $5\%$. Surprisingly, we see a slight performance drop for the KR-BERT model from the Rodong results. Further investigation of this performance drop might yield valuable insights. More importantly, we see that the DPRK-BERT model suffers from catastrophic forgetting. The well-learned representations that would enable $~51\%$ MLM accuracy are forgotten during the training on the DPRK dataset. We see that the model's performance drops to $25.788\%$, well below all the other PLMs. This strongly suggests that excessive fine-tuning without regularization causes catastrophic forgetting. These results also suggest that the datasets for the two languages are significantly different and the representations learned on them does not help each other. Compared to the DPRK-BERT model, CL-DPRK-BERT achieves significantly better results on the KorNLI dataset. In fact, CL-DPRK-BERT's $45.317$ accuracy with $\lambda=0.2$, outperforms mBERT by more than $3\%$ and achieves comparable results with the KR-BERT-MEDIUM model, which is trained only on the RoK Korean sentences. This shows the with cross-lingual regularization, the model successfully retains the well-learned RoK Korean representations.

Overall, when averaged over the two datasets (Rodong and KorNLI), we see that using the proposed cross-lingual regularization yields significant improvement on both evaluation metrics. With $\lambda=0.2$, wee see that log-perplexity is $0.311$ better than the second best model (KR-BERT). Moreover, the cross-lingual DPRK-BERT model outperforms the compared PLMs by more than $4.8\%$ MLM accuracy. These results demonstrate the usefulness of applying the cross-lingual regularizer towards achieving cross-lingual PLMs.

\subsection{New Year Speech Results}

Finally, we evaluate all models on the New year speeches. The results for both \textit{Perplexity} and \textit{MLM Accuracy} are given in Table~\ref{tab:mlm-newyear}. Interestingly, we see that the performance of all compared models increase when evaluated on New Year Speeches. For example, the average MLM accuracy of mBERT increases by $7.55\%$ (from $37.73\%$ to $45.28\%$). This strongly suggests that \textit{the use of Korean language in New Year speeches is closer to RoK Korean compared to Rodong news articles}. Conversely, \textit{Rodong news articles have a much more tailored and unique use of the Korean language}. Similar to the Rodong results, we observe that the DPRK-BERT model significantly outperforms all other approaches. $4.49$ improvement on perplexity and $16.72\%$ improvement on accuracy over the closest model (KR-BERT). We see a performance drop for DPRK-BERT  when evaluated on the New Year Speeches, $5.72\%$ drop in average MLM accuracy and $0.74$ increase in average perplexity. This is expected, as the DPRK-BERT is trained over the training split ($80\%$ of all dataset) of the Rodong news articles. Even though the training and validation splits are disjoint, the validation split naturally has a more similar data distribution to the training than New Year speeches. The cross-lingual DPRK-BERT model with $\lambda=0.7$ also significantly outperforms all compared PLMs except for the DPRK-BERT model. More importantly, unlike the DPRK-BERT model, the performance of the cross-lingual model on both DPRK datasets are comparable to each other (for both Rodong and New Year, it achieves around $69\%$ MLM accuracy). This shows that the cross-lingual model does not suffer from overfitting the Rodong training set.

% \begin{figure*}
   
%   \begin{subfigure}{0.5\textwidth}
%     \includegraphics[width=\linewidth]{materials/new-year-perplexity.png}
%     % \caption{Plot comparing average improvement in accuracy over STL with target dataset size.}
%     \caption{Perplexity}
%     \label{fig:newyear-perp}

% \end{subfigure}%
%  \begin{subfigure}{0.5\textwidth}
%     \includegraphics[width=\linewidth]{materials/new-year-accuracy.png}
%     % \caption{Plot comparing average improvement in accuracy over STL with target dataset size.}
%     \caption{Accuracy}
%     \label{fig:newyear-acc}
%     \end{subfigure}%

% \caption{Result for all four deep LMs on the New Year Speeches. The evaluation is repeated three times.}
% \label{fig:newyear-res}
% \end{figure*}

\begin{table}[ht]
    \centering
        \caption{MLM Results for all LMs  on New Year Speeches. }
    \label{tab:mlm-newyear}
\begin{tabular}{c|cc}\hline
Model&Perplexity&Accuracy\\\hline
mBERT&2.853&45.281\\
KR-BERT-MEDIUM&2.922&46.32\\
KR-BERT&2.009&59.933\\\hline
DPRK-BERT&1.088&76.652\\
CL-DPRK-BERT ($\lambda=0.7$)&1.426&68.884\\\hline
    \end{tabular}
\end{table}

For both DPRK datasets, the results clearly showed that the DPRK-BERT model performs significantly better than all other LMs. This suggests that the DPRK-BERT is a much better choice if we want to capture the subtle differences between the DPRK and the RoK Korean languages and be able to interpret the political agenda and intentions of Pyongyang more accurately. On average, we see that the cross-lingual DPRK-BERT model achieves the highest score. This shows that the cross-lingual regularization helps for better generalization across the different versions of the same Korean language, and suggests that the same approach for similar language pairs can yield improved results and better generalization. 

\subsubsection{Effect of $\lambda$ }
 
 An important point of consideration is to how much we want to regularize the model outputs. This is controlled by the $\lambda$ parameter in the loss function $\mathcal{L}_i$. At one extreme, setting $\lambda=0$ defaults to giving zero importance to preserving the initially learned representations. Here we present the results for different values of $\lambda$.

\begin{figure}
     \centering
     \includegraphics[width=\textwidth]{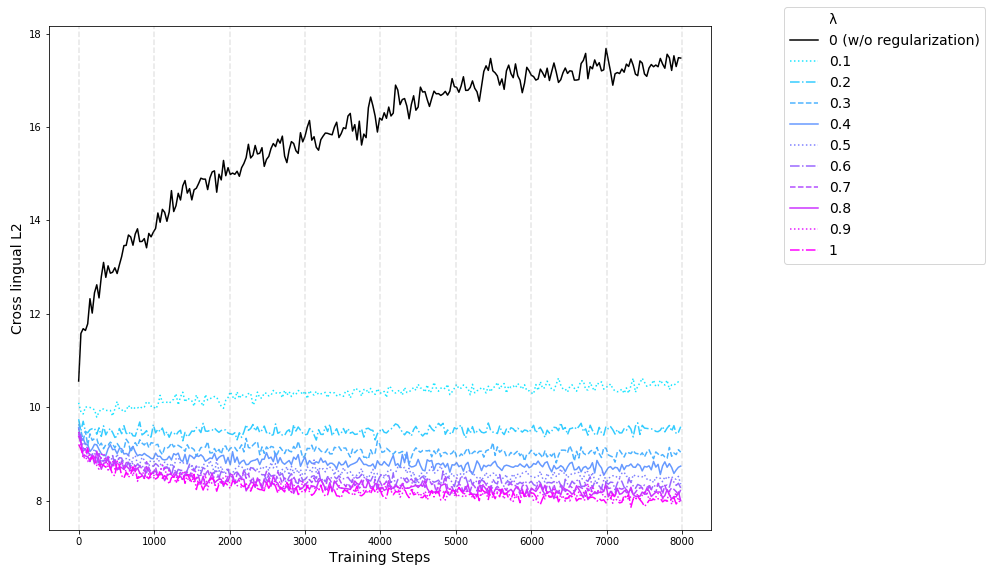}
     \caption{Effect of regularization on the representation stray during training.}
     \label{fig:crosslingual-ls}
 \end{figure}

 First, we analyze the stray of the representations from the initial ones during training. We refer to the stray as the cross-lingual L2 value. Figure~\ref{fig:crosslingual-ls} shows the results for different values of $\lambda$. We clearly see that the new representations stray rapidly without regularization. The representations learned with the regularization term do not stray from the initial weights. For all values of $\lambda$, the difference between the old and the new representations quickly converges. Models that use a higher valued $\lambda$ converges at a slightly smaller cross lingual L2. Overall the behavior of all cross-lingual models are similar.

\begin{figure*}
   
  \begin{subfigure}{0.5\textwidth}
    \includegraphics[width=\linewidth]{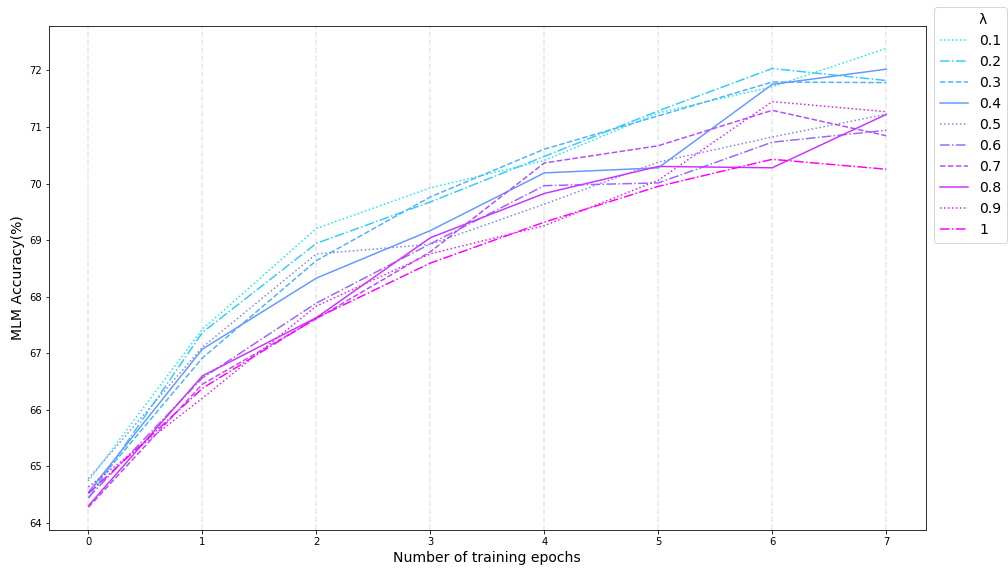}

    \caption{Rodong}
    \label{fig:perfclr-accrodong}

\end{subfigure}%
 \begin{subfigure}{0.5\textwidth}
    \includegraphics[width=\linewidth]{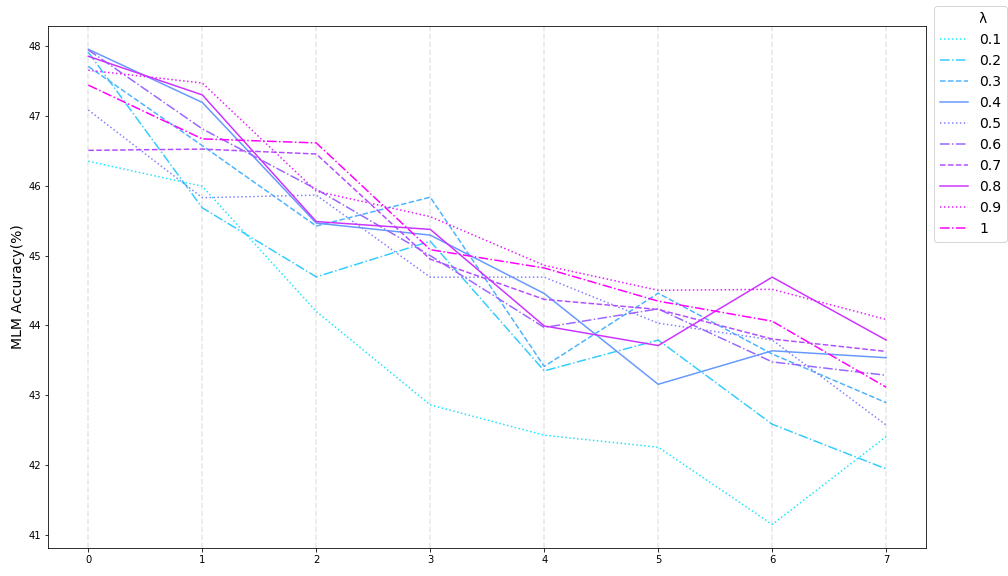}
    % \caption{Plot comparing average improvement in accuracy over STL with target dataset size.}
    \caption{KorNLI}
    \label{fig:perfclr-acckornli}
    \end{subfigure}%

\caption{MLM accuracy of cross-lingual models on the Rodong and KorNLI datasets during training.}
\label{fig:perfclr-acc}
\end{figure*}

\begin{figure*}
   
  \begin{subfigure}{0.5\textwidth}
    \includegraphics[width=\linewidth]{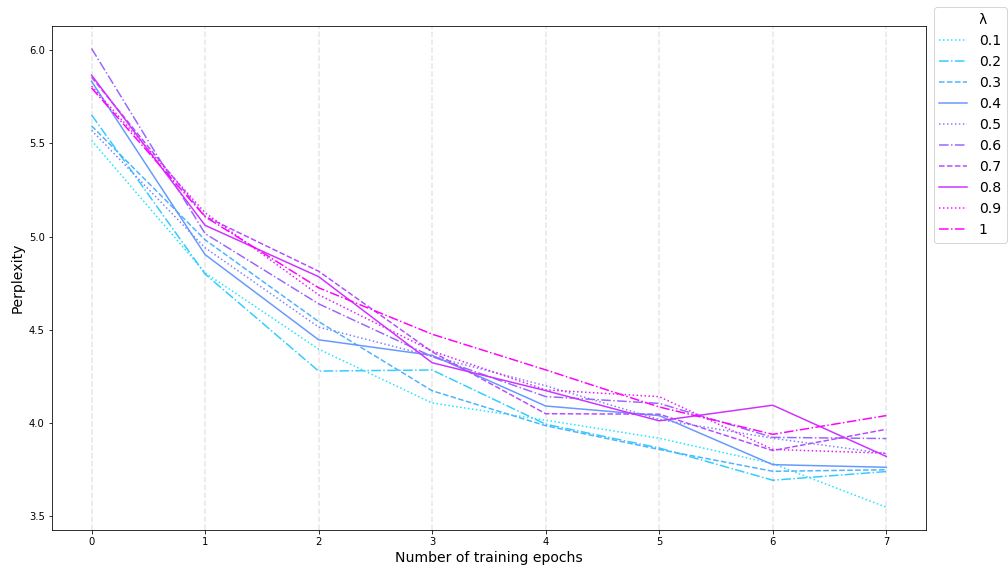}

    \caption{Rodong}
    \label{fig:perfclr-perprodong}

\end{subfigure}%
 \begin{subfigure}{0.5\textwidth}
    \includegraphics[width=\linewidth]{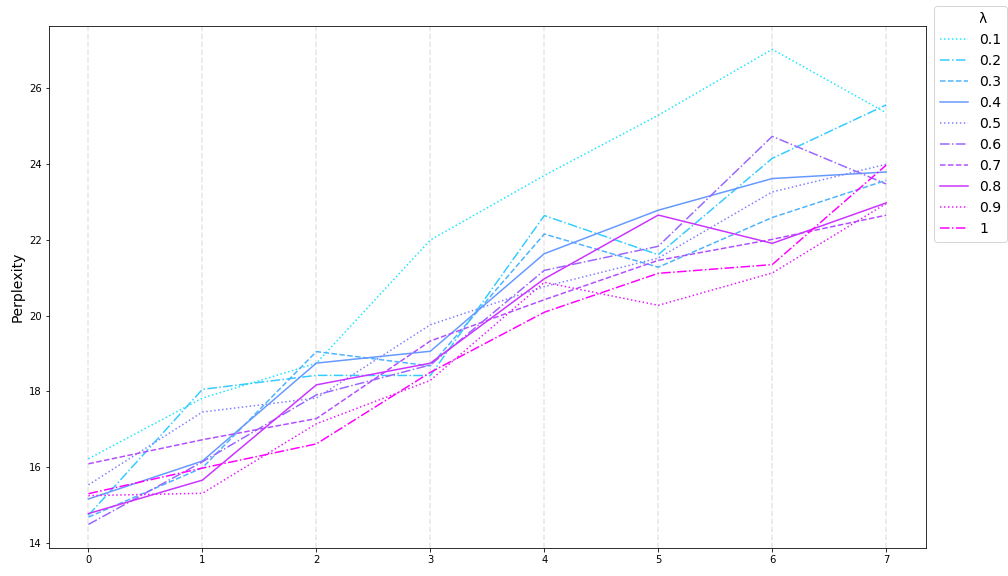}
    % \caption{Plot comparing average improvement in accuracy over STL with target dataset size.}
    \caption{KorNLI}
    \label{fig:perfclr-perpkornli}
    \end{subfigure}%

\caption{Perplexity of cross-lingual models on the Rodong and KorNLI datasets during training.}
\label{fig:perfclr-perp}
\end{figure*}

 Next, we analyze the performance on the Rodong and the KorNLI datasets during training for models that use a cross-lingual term. Figure~\ref{fig:perfclr-acc} and Figure~\ref{fig:perfclr-perp} show the MLM accuracy and the perplexity results, respectively. This analysis of useful to understand the behavior of the model with different values of $\lambda$. We see that for both metrics, the performance  changes on both datasets. For all values of $\lambda$, we see that the performance on the Rodong validation set improves and the performance on the KorNLI validation set  deteriorate. Especially, for the low values of $\lambda$, the performance on the KorNLI dataset deteriorates relatively quickly. Yet, for the large values of $\lambda$, e.g., $0.9$, the change in both datasets is much more gradual.

% \subsubsection{Error Analysis}

% Next, in order to better understand the source of error for the KR-BERT model, we manually analyzed its errors. The error analysis confirms that the systematic differences between the two Korean languages pose a challenge for the KR-BERT model. 
% [Table for examples where KR-BERT fails. Show that DPRK-specific language usage makes it challenging for KR-BERT and mBERT. Add examples that Yeojoo explained here...]

% If have time, focus on masking specific words and noun-phrases such as ideology phrases etc.

\section{Future Work}

In this paper, we developed the first DPRK Korean Language model, DPRK-BERT, by fine-tuning KR-BERT on Rodong news articles. Evaluated on two DPRK datasets, our DPRK-BERT language model significantly outperformed all other approaches. We compiled the first large-scale DPRK Korean dataset, around 30k news articles.~\footnote{Note that sharing it online as an open-source may have some legal issues. We will decide how to make this sharing process possible in consulting with the relevant parties.}

We also developed a new software to parse and download all Rodong news articles in text format. We believe that such NLP tools are valuable for fostering research on DPRK from non-CS disciplines such as Social Sciences.

In this study, we focused on the development of the first DPRK language model and showed significant improvement over the existing models. Next, we plan to apply this model for analyzing various DPRK texts such as new year addresses of the Supreme Leader(s) and other official statements. Also, we aim to capture the differences among the DPRK supreme leaders' signals by finding junctures among their new year addresses. Additionally, we pursue to compare the DPRK's signals to its domestic and international audiences by using the same texts in DPRK language and English with the DPRK-BERT and the current English language model. For instance, we may complement Lee et al.\cite{Lee2021UNSanc} by providing the comparison between the two language texts. We will employ topic modeling and sentiment analysis using DPRK-BERT on these texts and provide more meaningful results in the social science domain, such as the intentions and the signals of Pyongyang. 

We only considered a single method of introducing regularization. Different regularization methods might yield further improvements. Selectively regularizing the network parameters might yield further generalization~\cite{kirkpatrick2017overcoming}

\bibliographystyle{splncs04}
\bibliography{bib}

\appendix
\include{appendix}

\end{document}

%% file: appendix.tex
\section{Appendix}

\begin{table}[ht]
    \centering
        \caption{MLM Results for different values of $\lambda$. }
    \label{tab:mlm-rodongandkornliapp}
    \resizebox{\textwidth}{!}{\begin{tabular}{c|cc|cc|cc}\hline
    &\multicolumn{2}{c}{Rodong}  &\multicolumn{2}{c}{KorNLI}  &\multicolumn{2}{c}{Average}\\
$\lambda$&Perplexity&Accuracy&Perplexity&Accuracy&Perplexity&Accuracy\\\hline
0.1&1.492&68.704&2.951&44.927&2.222&56.816\\
0.2&1.495&68.429&2.903&45.317&\textbf{2.199}&56.873\\
0.3&1.448&69.716&2.989&45.067&2.218&\textbf{57.391}\\
0.4&1.614&66.939&2.836&46.247&2.225&56.593\\
0.5&1.515&68.492&2.913&45.587&2.214&57.04\\
0.6&1.629&66.425&2.838&45.949&2.234&56.187\\
0.7&1.439&69.481&3.038&44.425&2.238&56.953\\
0.8&1.639&66.287&2.773&47.192&2.206&56.74\\
0.9&1.636&66.324&2.802&46.932&2.219&56.628\\
1&1.555&67.697&2.913&45.227&2.234&56.462\\\hline
    \end{tabular}}
\end{table}

\begin{table}[ht]
    \centering
        \caption{MLM Results on the New Year Speeches for different values of $\lambda$. }
    \label{tab:mlm-newyear}
\begin{tabular}{c|cc}\hline\\
$\lambda$&Perplexity&Accuracy\\\hline
0.1&1.471&68.037\\
0.2&1.459&68.436\\
0.3&1.444&68.832\\
0.4&1.533&67.03\\
0.5&1.462&68.208\\
0.6&1.440&68.222\\
0.7&1.426&68.884\\
0.8&1.536&66.902\\
0.9&1.538&66.926\\
1&1.511&67.364\\\hline
    \end{tabular}
\end{table}

\begin{table}[ht]
    \centering
        \caption{Result for MLM on Rodong (DPRK) and KorNLI (RoK) datasets. }
    \label{tab:mlm-rodongandkornli}
    \resizebox{\textwidth}{!}{\begin{tabular}{c|cc|cc|cc}\hline
    &\multicolumn{2}{c}{Rodong}  &\multicolumn{2}{c}{KorNLI}  &\multicolumn{2}{c}{Average}\\
Model&Perplexity&Accuracy&Perplexity&Accuracy&Perplexity&Accuracy\\\hline
mBERT&3.41&37.734&3.066&42.084&3.238&39.909\\
KR-BERT-MEDIUM&3.241&43.129&2.928&45.553&3.085&44.341\\
KR-BERT&2.505&53.57&2.536&51.426&2.52&52.498\\
DPRK-BERT&0.803&82.367&5.465&25.788&3.134&54.078\\\hline
    \end{tabular}}

\end{table}